\documentclass[10pt,twocolumn,letterpaper]{article}

\usepackage{wacv}
\usepackage{times}
\usepackage{epsfig}
\usepackage{graphicx}
\usepackage{amsmath}
\usepackage{amssymb}
\usepackage{multirow}
\usepackage{program}
\usepackage{enumitem}
\usepackage{url}

\newcommand{\acro}{M-PACT}

\wacvfinalcopy 

\def\wacvPaperID{350} 

\ifwacvfinal\fi
\begin{document}

\title{M-PACT: An Open Source Platform for \\ Repeatable Activity Classification Research}

\author{Eric Hofesmann \\
CSE, University of Michigan\\
{\tt\small erichof@umich.edu}
\and
Madan Ravi Ganesh \\
ECE, University of Michigan\\
{\tt\small madantrg@umich.edu}
\and
Jason J. Corso \\
ECE, University of Michigan\\
{\tt\small jjcorso@umich.edu}
}

\maketitle
\ifwacvfinal\thispagestyle{empty}\fi

\begin{abstract}
There are many hurdles that prevent the replication of existing work which hinders the development of new activity classification models.
These hurdles include switching between multiple deep learning libraries and the development of boilerplate experimental pipelines.
We present \acro~to overcome existing issues by removing the need to develop boilerplate code which allows users to quickly prototype action classification models while leveraging existing state-of-the-art (SOTA) models available in the platform.
\acro~is the first to offer four SOTA activity classification models, I3D, C3D, ResNet50+LSTM, and TSN, under a single platform with reproducible competitive results.
This platform allows for the generation of models and results over activity recognition datasets through the use of modular code, various preprocessing and neural network layers, and seamless data flow.
In this paper, we present the system architecture, detail the functions of various modules, and describe the basic tools to develop a new model in \acro.

\end{abstract}


\section{Introduction}
\label{sec:intro}

Empirical fields like activity classification in computer vision demand repeatable and measurable software systems to support benchmarking and research progress.
The availability of large-scale computational resources as well as open-source deep learning software, including Tensorflow~\cite{abadi2016tensorflow}, Pytorch~\cite{paszke2017automatic}, and Caffe~\cite{jia2014caffe}, have taken steps toward providing such software systems.  
Shared, open-source deep learning library code reduces the expected time to develop new models.
However, these libraries fall short of meeting the full demands of repeatable and measurable software systems:  models developed in one library are not easily integrated with models from another library; reported results may not use the same practices or even be reproducible; the necessity to switch between different libraries greatly slows down the research cycle.
The delays in model integration across libraries are, in part, due to differences in compilation procedures and general coding structures wherein even keywords can significantly vary in meaning.
For example, ``\texttt{dropout}'' can indicate the ratio of nodes to keep (Tensorflow) or to drop (PyTorch), depending on the library being used.

\begin{figure}[t!]
    \begin{center}
\includegraphics[width=\columnwidth]{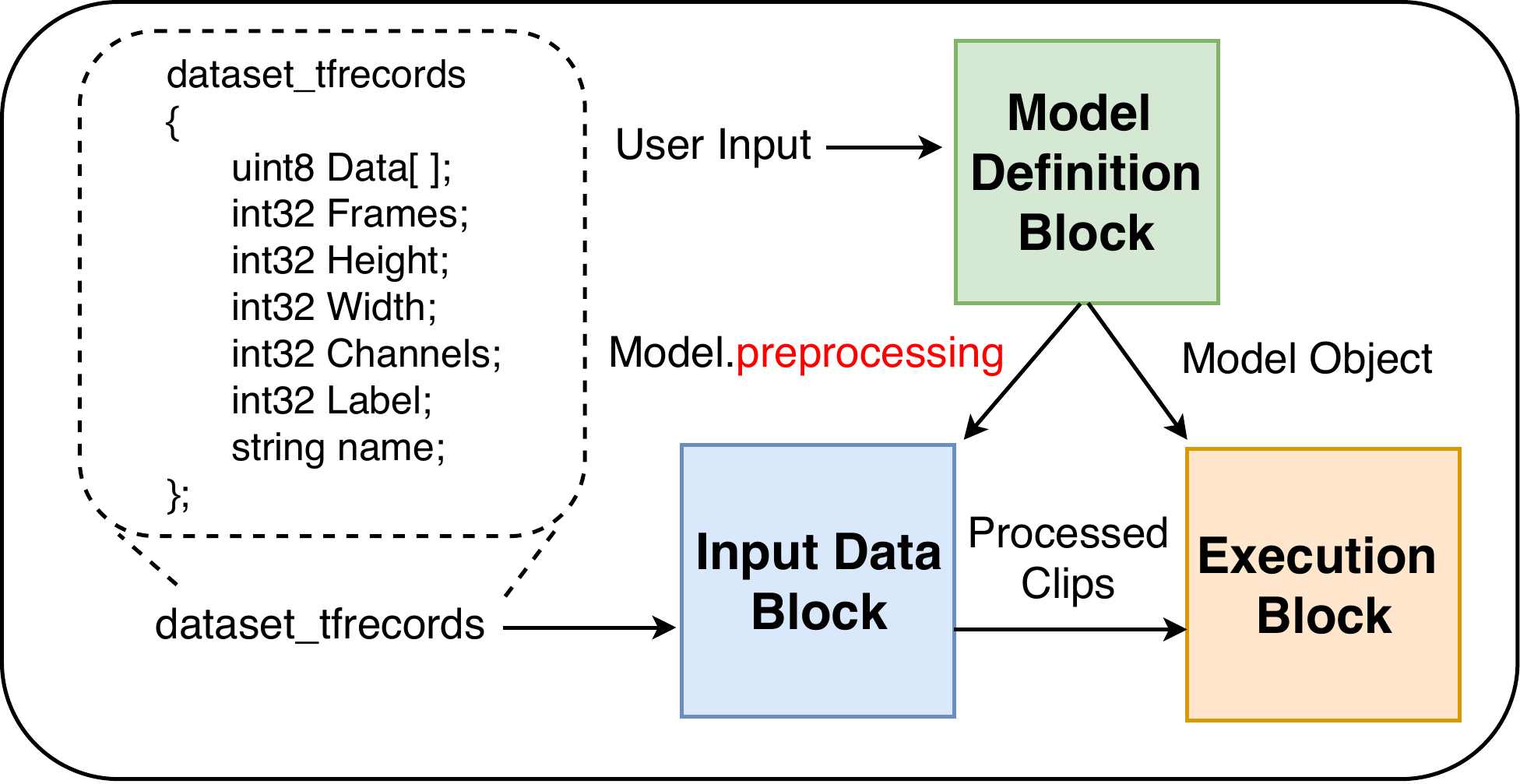}
    \end{center}
\caption{Illustration of the three main components of \acro~and how they interact with each other.}
\label{fig:overview}
\end{figure}

In this work, we introduce the Michigan Platform for Activity Classification in Tensorflow, a platform that allows the user to focus solely on the creation or fine-tuning of models while providing simple abstractions for functions like data input, metric computation, and feature extraction.
Users do not need to build their experimental pipeline from scratch and can quickly prototype models to run experiments.
Furthermore, \acro~provides implementations of several SOTA models allowing users to leverage the potential of existing models.
Since \acro~is developed using a single language and library, Python and Tensorflow, and a consistent coding style throughout, it avoids the complication of interpreting code written in different libraries or following different practices.
Thus, \acro~removes the necessity of re-implementing existing code from other libraries and the time to set up and run experiments all while keeping the user focused on developing new activity classification models.

\begin{figure*}[t!]
    \begin{center}
\includegraphics[width=2\columnwidth]{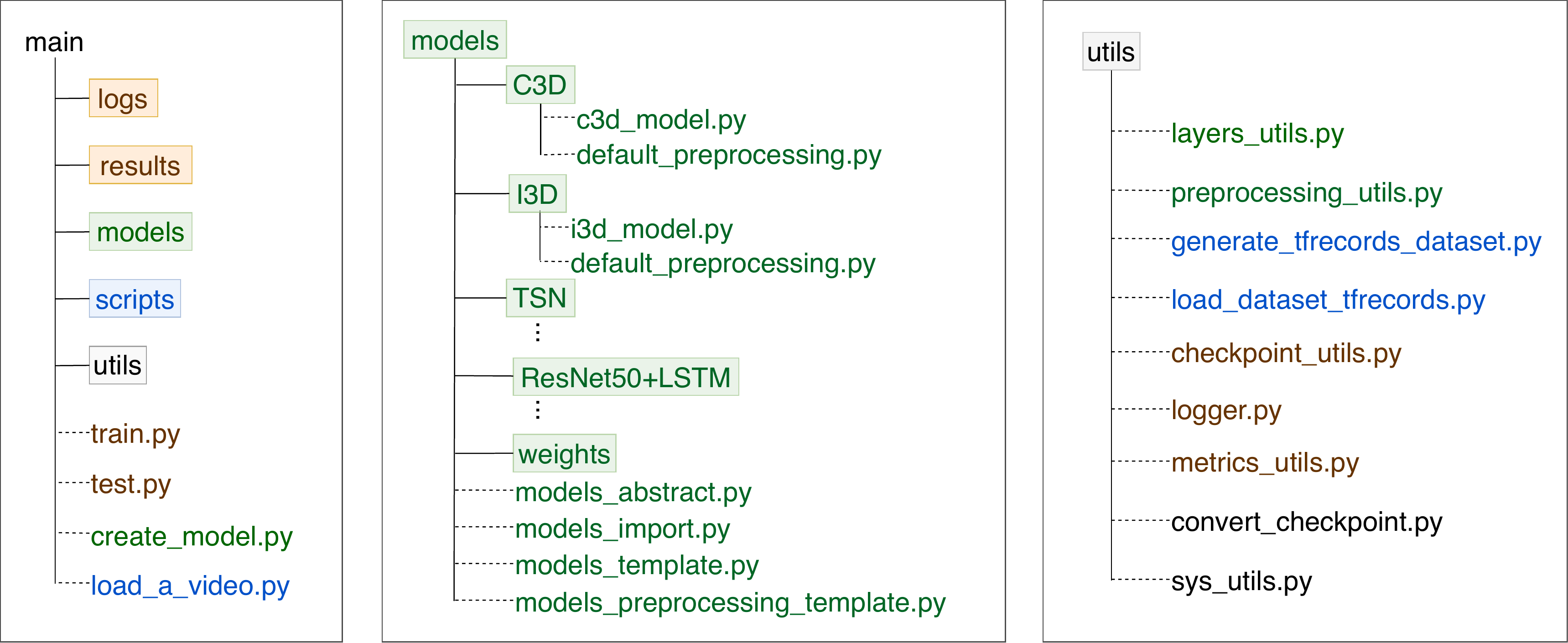}
    \end{center}
\caption{
Illustration of the \acro~directory structure.
Directories are displayed with boxes around their names and solid lines.
Files are illustrated with only dotted lines and ending in ``.py''. 
    From left to right, the figure shows the contents of the root directory and then two of the most important directories, \texttt{models} and \texttt{utils}.
}
\label{fig:file_structure}
\end{figure*}

The main contributions of \acro~are:
\begin{itemize}[itemsep=0ex,topsep=0ex,parsep=0pt]
\item{In terms of usage}
\begin{itemize}[itemsep=0ex,topsep=0ex,parsep=0pt]
\item Availability of SOTA activity classification models under a common platform.
\item A multitude of metrics that allow for quick and efficient model testing which standardizes comparisons among various models.
\item Ability to extract features from existing SOTA activity classification models.
\item Single node multi-GPU support to train and fine-tune models.
\end{itemize}
\item{In terms of design}
\begin{itemize}[itemsep=0ex,topsep=0ex,parsep=0pt]
\item Efficient and clear coding style, supplemented with simple comments, that allows users to assimilate and develop models quickly.
\item Modular code design that offers a vast array of customizable options for deep neural network layers and video preprocessing functions.
\item Customized input pipeline that condenses the loading and processing of videos into a set of flexible and easy-to-use options for the user.
\end{itemize}
\end{itemize}



\section{Related Work}
\label{sec:relworks}

Most deep learning libraries, like Caffe, have a built-in model zoo containing standard image feature extraction models like VGG16~\cite{Simonyan14verydeep}, AlexNet~\cite{krizhevsky2012imagenet}, and ResNet~\cite{he2016deep} to help bootstrap model development.
To the best of our knowledge, these libraries do not offer models pre-trained on video-based tasks.
Within most standard libraries, videos are accessed and handled explicitly as a collection of image files under a specific video directory.
This introduces large processing overheads with respect to the file access-read-process functionality.
When combined with the lack of video- and clip-specific preprocessing functions, this forces video-based models to perform suboptimally.

To the best of our knowledge we are the first to provide a compilation of SOTA activity classification algorithms under a single library.
This is a difficult task because of two reasons, 1) cross-compatibility issues between different versions of a library and 2) differences in function implementations between libraries.
For example, in Caffe, researchers modify the base code causing cross-compatibility issues when combining multiple models.
In Tensorflow, variants like TF-slim~\cite{silbermantf} and Keras~\cite{chollet2015keras} form unreliable bases for \acro~because of subtle differences in function implementations and varying abstraction levels.

Among the many possibilities of contemporary libraries, we chose Tensorflow since it is well-adapted and known to produce fast systems.
Through Tensorflow, \acro~can provide video- and clip-level preprocessing functions along with existing frame-wise preprocessing.
By explicitly dealing with data as videos and clips, offering a vast collection of clip-level functions to the user, and combining some of the best features from existing libraries we provide a large platform for the quick development of video-based models.


\section{Breakdown of \acro}
\label{sec:breakdown}

\acro~consists of three main components,  1) \textbf{Input Data Block}, 2) \textbf{Model Definition Block}, and 3) \textbf{Execution Block}, as shown in Fig.~\ref{fig:overview}.
Aside from custom tasks, the \textbf{Model Definition Block} is the only portion of \acro~in which the user is required to write code.


\acro~follows a strict file structure convention, shown in Fig.~\ref{fig:file_structure}, that must be followed.
The main \acro~directory contains the python files necessary to train, test, and create a model and the directories containing \texttt{logs}, \texttt{results}, \texttt{models}, \texttt{scripts}, and \texttt{utils}.
\texttt{Results} and \texttt{logs} contain the model weights and tensorboard logs of trained and tested models.
\texttt{Scripts} contains the shell script used to download the initial weights for the provided models in \acro.
The \texttt{utils} and \texttt{models} directories store a host of useful files and directories with a fixed structure.

\acro~provides extensive flexibility in loading data and quickly prototying and experimenting on new models.
The following sections will provide a deeper discussion about the \textbf{Input Data Block}, the \textbf{Model Definition Block}, and the \textbf{Execution Block} respectively.



\subsection{Input Data Block}
\label{subsec:ipblock}

\begin{figure}[t!]
    \begin{center}
\includegraphics[width=\columnwidth]{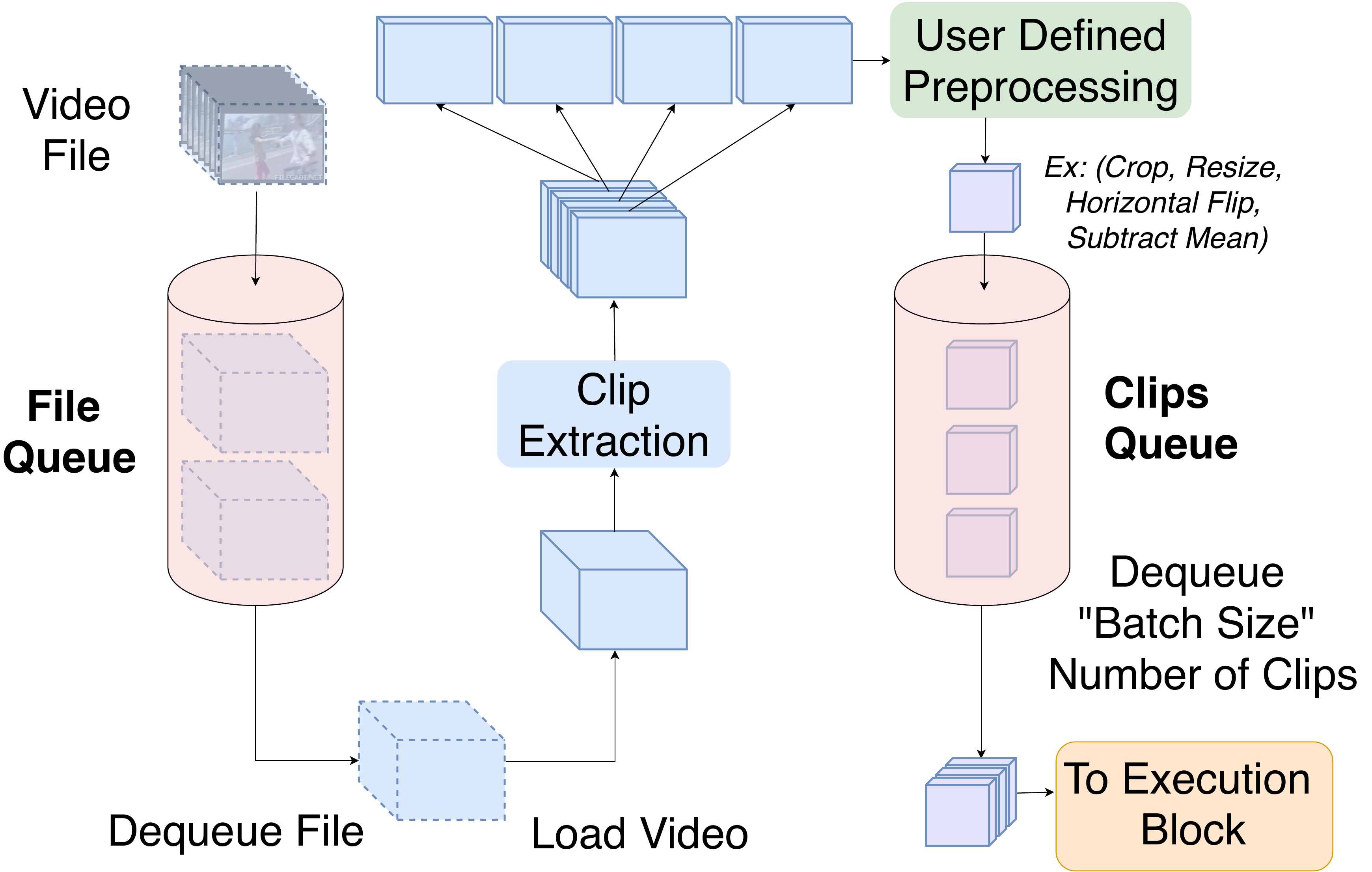}
    \end{center}
\caption{The \textbf{Input Data Block} consists of two queues, the file and clips queue.
The entire input data flow is illustrated in this figure.
    This includes 1) loading video filenames into the queue to breaking them down into clips, 2) preprocessing each clip individually, and 3) using the clips queue to dequeue them and provide mini-batches of data to a model.}
\label{fig:ip_block}
\end{figure}

The \textbf{Input Data Block} is the entry point for data within \acro.
Its overall functionality can be divided into two broad stages: 1) reading video data, from TFRecords, into a format compatible with the platform, and 2) extracting clips from videos and preprocessing them.
Fig.~\ref{fig:ip_block} illustrates the structure and flow of data as it passes through this block.

\subsubsection{Read video data}
\label{subsubsec:readdata}

All the components of the \textbf{Input Data Block} are constructed in Tensorflow to facilitate efficient and parallel data loading.
The names of the video files for a selected dataset are loaded into a file queue.
Each dequeued file name is then read and broken down into a user-specified number of clips which get processed by a model-specific preprocessing function.
To allow the extraction of multiple clips from a given video, a separate clips queue is used to store all of the clips.
The file queue and clips queue work together to ensure exactly one mini-batch worth of clips is passed into a model during each iteration.
If a video contains more than one mini-batch of clips, then the excess clips will be provided to the next mini-batch.
When there are too few clips in a video to form a mini-batch, then the missing clips are generated from a new video.

\begin{figure}[t!]
    \begin{center}
\includegraphics[width=\columnwidth]{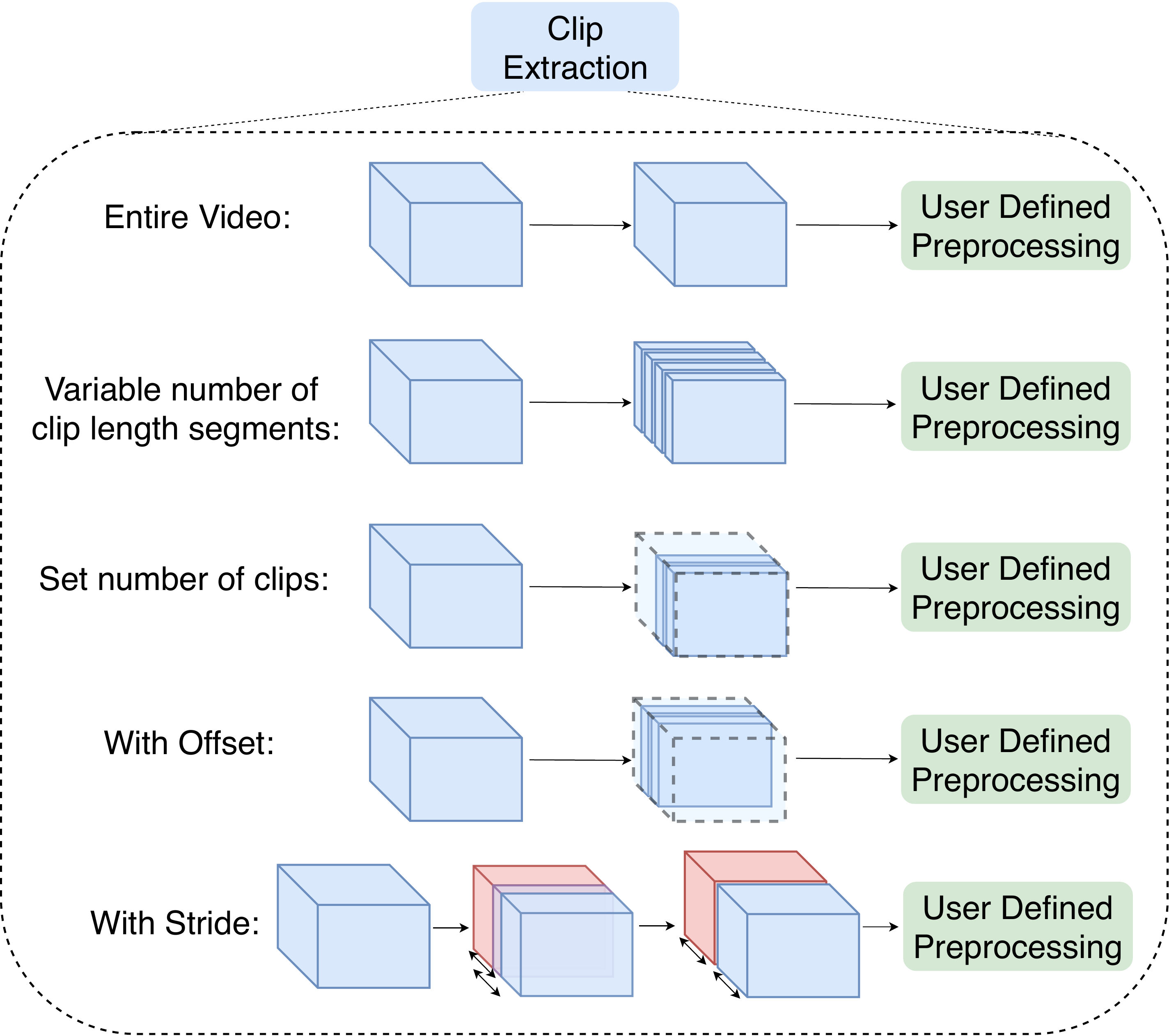}
    \end{center}
\caption{
This figure shows the various ways that a video can be broken into clips.
By default the entire video is passed directly in to a model-specific preprocessing function without any modifications.
The length and number of clips are defined by command line arguments.
Clip offset can be used to select clips at a point other than the beginning of the video while clip stride defines the amount of overlapping frames between different clips.
A positive stride value indicates a space between clips while a negative value indicates the amount of overlap.
In addition to the arguments shown, \acro~also provides an option for random clip selection.
}
\label{fig:extract_clips}
\end{figure}

\begin{figure*}[t!]
    \begin{center}
\includegraphics[width=2\columnwidth]{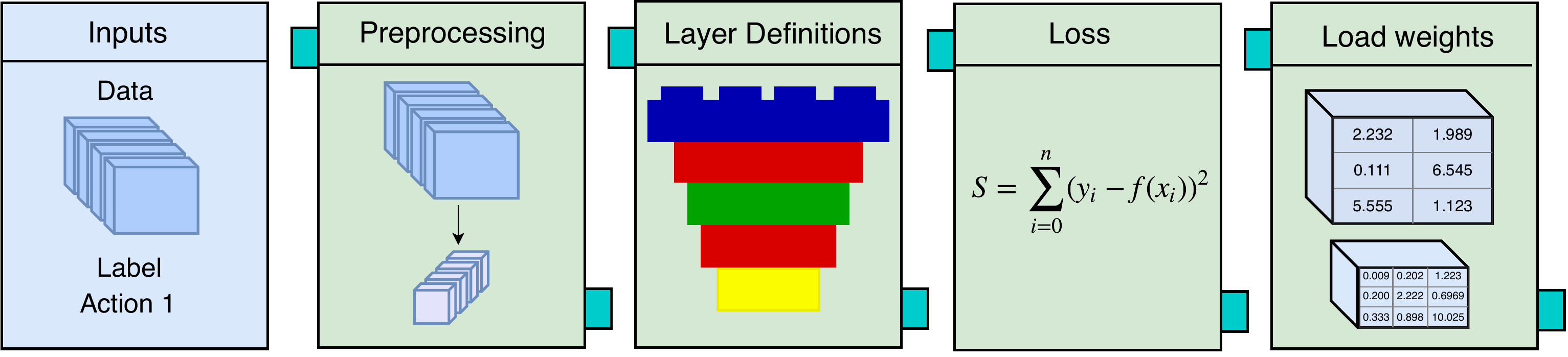}
    \end{center}
\caption{
The \textbf{Model Definition Block} is comprised of the 1) input video preprocessing steps, 2) network layer definitions, 3) loss specification, and 4) initial weights to load.
This block contains the aspects of \acro~that the user develops when creating a new model.
The leftmost ``Inputs'' section of this figure represents the video data that is passed from the \textbf{Input Data Block} to the \textbf{Model Definition Block}.
}
\label{fig:model_block}
\end{figure*}

\subsubsection{Extract clips}
\label{subsubsec:extractclips}

In typical video understanding work, it is sometimes more useful to divide a video into multiple clips and process them individually instead of operating on the entire video at once. 
To this end, \acro~offers multiple options to break down a video into any desired number of clips. 
In order to do so, a number of clip specifications such as the clip length, total number of clips per video, offset from beginning of video to select a clip, and the stride of the sliding window from which clips are extracted, can be provided in any permutation.
These specifications are provided in the form of command line arguments.
Fig.~\ref{fig:extract_clips} lists a number of possible ways to construct a clip(s) from a given video.

Once clips have been extracted, they are automatically processed using a selected model's preprocessing pipeline which is defined within the \textbf{Model Definition Block}.
Preprocessed clips can significantly vary in shape (height and width) as well as the number of frames when compared to the input video.
However, the shape of the preprocessed clip is fixed throughout the remainder of the pipeline and each subsequent block regardless of the shape of the input video.


\subsection{Model Definition Block}
\label{subsec:model_def}

The \textbf{Model Definition Block}, shown in Fig.~\ref{fig:model_block}, includes the files and procedures that form a model in \acro~and it is the only portion which requires mandatory coding.
This block contains user-defined input video preprocessing steps, network architecture and layer definitions, losses associated with a model, and pre-trained initialization weights.
To ease model development, \acro~contains pre-defined preprocessing and network layer definitions, as well as an automated template generation system to ensure compatibility of a new model with the rest of the platform.
Additionally, a multitude of command line arguments, for example the \texttt{sequence\_length} of an LSTM, are also accessible within the model class.

\subsubsection{Preprocessing}
\label{subsubsec:preprocessing}

The preprocessing pipeline, a crucial component of the \textbf{Input Data Block}, is defined with the \textbf{Model Definition Block}.
There is a host of video and image processing steps which are commonly used for activity classification.
These steps exist to either standardize the frames between videos, so that the input to the network is consistent, or to augment input data to alter the way the network learns.
\acro~offers a collection of common preprocessing functions, e.g. cropping and resizing, to be used in developing a custom preprocessing pipeline.
More importantly, \acro~is the first, to our knowledge, to offer video- and clip-level preprocessing functionality like temporal resampling, clip-based cropping or shuffling, and more.

With the wide range of preprocessing options that are available, it is important to try different combinations to determine which help a model perform well.
To facilitate this, \acro~supports the implementation of multiple preprocessing pipelines for a single model which can be selected using a command line argument during training or testing time.
Each of these preprocessing pipelines must exist in independent files wherein each file contains a \texttt{preprocess} function which returns a processed video.
This \texttt{preprocess} function takes a clip from the \textbf{Input Data Block} and returns the clip processed both spatially and temporally.
The shape of this processed clip must match that expected by the model.

\subsubsection{Model Architecture}
\label{subsubsec:modelarch}

Once the shape and characteristics of the processed videos have been fixed, they are passed in to the \texttt{inference} function of the model.
The \texttt{inference} function contains the entire sequential definition of the layers that make up the model.
\acro~offers a collection of pre-defined neural network layer definitions, similar to the collection of functions available for video preprocessing, including 2D and 3D convolutions, fully connected layers, and more.
These layers also occasionally add new functionality that is not available in Tensorflow; e.g., the grouping of convolutional filters from Caffe and custom height and width padding.
Since \acro~is built atop Tensorflow, any basic Tensorflow functions are also applicable between layers.

The model file itself contains the definition of the model class and relevent methods including inference, loss, and initial weight loading, all of which are unique to each model. 
As previously mentioned, the preprocessing functions are defined in separate files and called inside of the model file while the inference, loss, and loading of initial weights are defined directly in the model file.
The loss function is internally defined within each model in order to offer the flexibility of customized losses.
Similar to preprocessing, the user may want to test various losses to determine which is most useful for a given model.
\acro~contains a command line argument that allows the user to switch between different user-defined loss types.
Finally, the model class also contains an option to load initial weights for a model that has been pre-trained within the platform or otherwise.


\subsection{Execution Block}
\label{subsec:exec}

\begin{figure}[t!]
    \begin{center}
\includegraphics[width=\columnwidth]{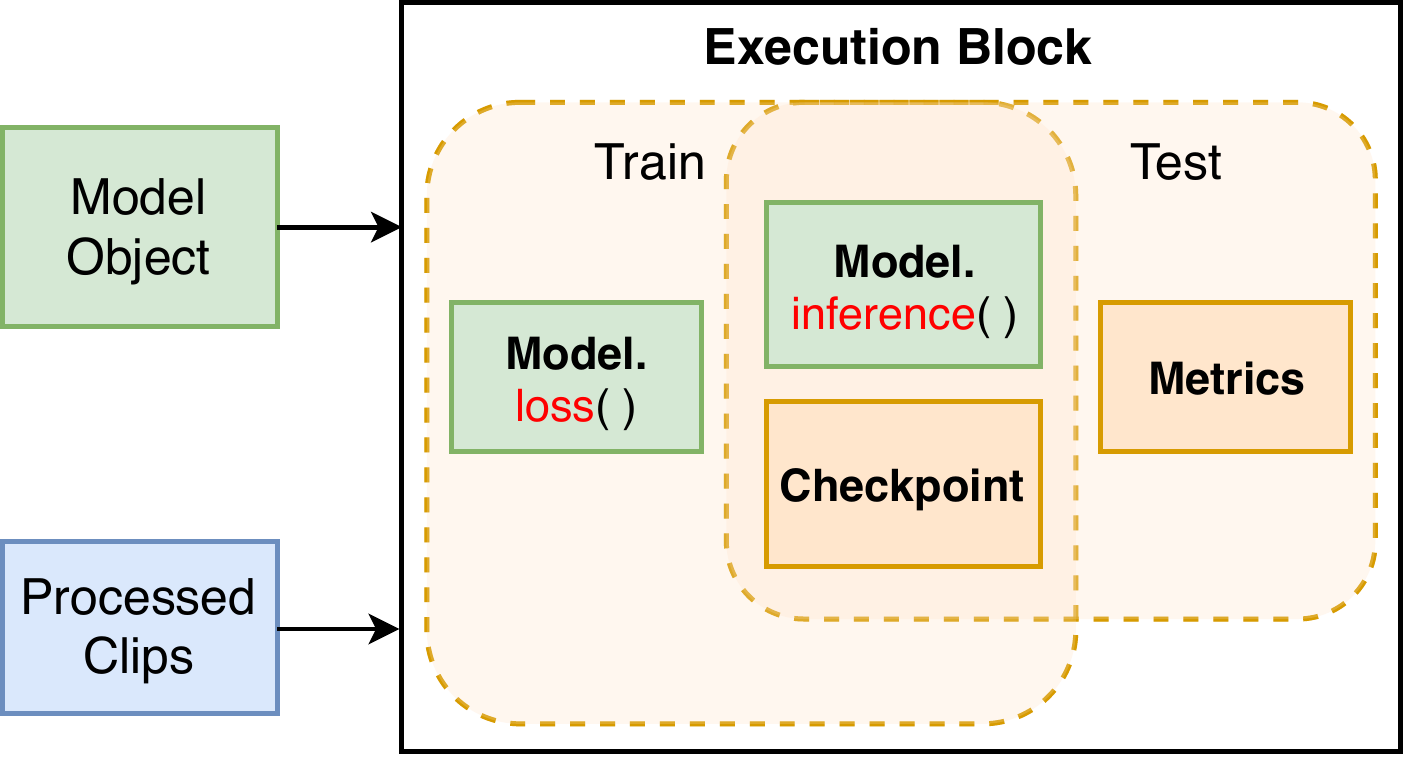}
    \end{center}
\caption{Training and testing are the two major phases within the execution block.
Within these phases, the checkpoint and metrics modules save model weights and calculate performance metrics.
Between these modules, checkpoint-based functions are part of both phases while metrics are only calculated during testing.}
\label{fig:exec_block}
\end{figure}

The training and testing of activity classification models requires boilerplate code that every researcher must write before being able to run experiments.
The \textbf{Execution Block} abstracts away this boilerplate code to a set of flexible command line arguments while at the same time offering the added functionality of checkpoint and metrics modules.
Fig.~\ref{fig:exec_block} illustrates the general outline of the \textbf{Execution Block} and highlights the key modules used in training and testing phases.
The following sections detail the relevant functionality of the \textbf{Execution Block} and describe the checkpoint and metric calculation systems of \acro.

\subsubsection{Training}
\label{subsubsec:training}

The training pipeline used in \acro~is similar to most standard Tensorflow-based training pipelines.
It includes the definition of a model and an optimizer, loading of weights, application of gradients, execution of a Tensorflow session that runs the training operation, and the saving of a model's state and metrics.

Two important points to note in the training process of \acro~are the utilization of GPUs and the definition of epochs.
Once the setup is complete, the model is replicated across a selected number of GPUs in the compute node and processed clip data from the \textbf{Input Data Block} are interfaced with the model.
Currently, \acro~only supports the extension of a model to multiple GPUs within a single compute node.

During the execution of the training phase, models are trained for a user-specified number of epochs.
The number of iterations in each epoch is based on the number of videos in a given dataset, the mini-batch size, and the number of clips per video.
Within a user-specified frequency of epochs, \acro~saves a model's state using the \textbf{Checkpoint Module}.\texttt{save()} function.
It is important to note that during the setup of an experiment, \acro~offers adaptive learning rate control which steps down the learning rate when the training loss plateaus, alongside a variety of native optimizers.

\subsubsection{Testing}
\label{subsubsec:testing}

The testing pipeline loads the most recently saved state of a selected model and can be used to extract features or to calculate a model's classification accuracy using the \textbf{Metrics Module}.
When extracting information from a model, any layer within the specified model can be chosen using a command line argument. 
Additional command line arguments are available during the testing phase which add functionality like averaging predictions across clips of a video instead of classifying each clip individually or applying a softmax function to model outputs for classification.

\subsubsection{Checkpoint Module}
\label{subsubsec:checkpoint}

An important functionality in any experimental setup is ability to save and load the state of a model in the form of checkpoints.
The \textbf{Checkpoint Module} contains functions that handle all of the essential operations with respect to checkpoints.
The ability to save and recover checkpoints is crucial to postprocess data, fine-tune a model from any desired state, and restart training in case of a critical hardware failure.
Each checkpoint is saved in a unique directory corresponding to the chosen dataset, preprocessing method, experiment name, and metric.

It is often difficult to convert checkpoints between various formats.
For example, when weights from a Caffe model need to be loaded into a Tensorflow model, the \texttt{.caffemodel} must be loaded in Caffe and saved, possibly as a \texttt{numpy}~\cite{oliphant2006guide} file, before starting a Tensorflow session and saving the weights as a \texttt{.ckpt} file.
To avoid this two-step process, the \textbf{Checkpoint Module} automatically saves and loads weights in a \texttt{numpy} format.
This allows weights generated using \acro~to be loaded more easily into other deep learning libraries.
However, loading weights trained in other deep learning libraries into \acro~is more complicated due to the variability in how model weights are stored and their dimensionality between libraries.
Since there is no one-size-fits-all approach to converting weights into \acro, we provide a detailed code example in \texttt{utils\textbackslash convert\_checkpoint.py} to help guide the user in converting model weights to an \acro-compatible format.


\subsubsection{Metrics Module}
\label{subsubsec:metrics}

Evaluating the performance of a trained model is a key part of the testing pipeline.
\acro~offers the \textbf{Metrics Module} for testing and logging the activity classification performance of a given model as well as extracting features from it.
During testing, the chosen metric and any other specified variable is logged automatically using \texttt{tensorboard}.

Available performance metrics include average pooling and last frame prediction with the option of training a linear SVM classifier on extracted features.
Average pooling is a common technique which uses the softmax per-class predictions of a model and averages them across all frames of an input video.
From these averaged predictions, the label with the maximum value is used as the overall class prediction for that video.
Last frame prediction simply uses the softmax per-class prediction of the last frame in the video to generate the overall class prediction.
Last frame classification is referenced originally in the ConvNet+LSTM model~\cite{donahue2015long}.
In addition to these metrics, which directly classify the output of a model, \acro~also provides the option of training a linear SVM classifier based off of features extracted from any user-specified layer within the model.
This SVM is tested within the \textbf{Metrics Module} using the features extracted from each testing video.

The method used to calculate the performance of a model can vary depending on the output of the model.
For example, C3D classifies multiple clips in a video individually and then averages these predictions to produce a single video-level prediction.
On the other hand, TSN produces a single prediction for each video.
The \textbf{Metrics Module} is robust enough to handle both such cases.


\begin{table*}[t!]
    \begin{center}
        \begin{tabular}{c|c|c|c|c||c|c|c}
            \hline
            \multirow{2}{*}{Base Model} & \multirow{2}{*}{Pretraining} & \multicolumn{3}{c||}{HMDB51 Acc. (\%)} & \multicolumn{3}{c}{UCF101 Acc. (\%)} \\\cline{3-8}
            & & Ours & Authors & $\Delta$ & Ours & Authors & $\Delta$ \\
            \hline
            C3D & Sports-1M & 52.94 & 50.30 * & +2.64 & 79.14 & 82.30 * & -3.16 \\
            I3D & Kinetics & 66.54 & 74.80 * & -8.26 & 93.18 & 95.60 * & -2.42 \\
            TSN & ImageNet & 51.70 & 54.40 & -2.70 & 85.25 & 85.50 & -0.25 \\
            ResNet50+LSTM & ImageNet & 45.36 & 43.90 & +1.46 & 79.25 & 84.30 & -5.05 \\
            \hline
        \end{tabular}
   \end{center}
        \caption{Mean RGB classification accuracies are shown for various SOTA action classification models across split 1 of HMDB51 and UCF101.
        Values marked with a (*) indicate averaged results across all three splits of a given model.
        Accuracies are reported for each baseline model and pretraining dataset combination.
        The weights used were pretrained on the video recognition datasets Sports-1M~\cite{KarpathyCVPR14} and Kinetics~\cite{he2016deep} as well as the image recognition dataset ImageNet~\cite{deng2009imagenet}.}
        \label{tab:model_results}
\end{table*}


\section{Using M-PACT}
\label{sec:workflow}

\acro~is developed around two main use cases: 1) to extract features from activity classification models implemented in \acro, either as video representations or for classification, and 2) to serve as a development platform for new activity classification models.
The only time that the user is required to write any code is when setting up a new model's layer architecture and preprocessing pipeline.
However, in order to successfully use the platform, the user must first convert their video dataset into the format expected by \acro.

\subsection{Dataset Conversion}
\acro~uses TFRecords to store videos and Tensorflow queues to parse the data.
The efficiency of the \acro~pipeline comes from the fact that the reading and processing of data is built into the Tensorflow graph and no external user input is required.

Any video dataset needs to converted to TFRecords in order to be compatible with the \acro~data pipeline.
This can be accomplished by executing \texttt{generate\_tfrecords\_dataset.py}, found under the \texttt{utils} directory as shown in Fig.~\ref{fig:file_structure}.
This conversion file expects data to be stored in a video format, e.g. avi or mp4, within subdirectories named according to the video's action class.
The input to the conversion file is a command line argument indicating the path to the parent directory of the dataset.
The final TFRecord, one for each video, stores eight attributes: data, number of frames, height, width, number of channels, true label, and the name of the video file as shown on the left side of Fig.~\ref{fig:overview}.

\subsection{Existing Models}
Once a dataset is converted to the required format, users can readily apply the existing SOTA activity classification models, I3D, C3D, ResNet50+LSTM, and TSN, to any custom dataset.
Since the SOTA models are available within \acro, users can quickly compare with or improve upon the generated results.
The complete model definition and relevant weights, including a script to retrieve the weights from an online server, is included within the platform.
Classification results or features can be extracted from these models by running \texttt{test.py} along with relevant command line arguments.
If a dataset other than HMDB51~\cite{kuehne2013hmdb51} or UCF101~\cite{soomro2012ucf101} is used to generate results on existing models, then these models will need to be fine-tuned by running \texttt{train.py} with relevant command line arguments.

\subsection{Model Development}
\acro~streamlines the training and testing pipeline to run from simple command line arguments and also requires only a minimal amount of coding to develop and use a new model.
Model development begins by using the \texttt{create\_model.py} program to generate a template for the model and preprocessing files, as shown in Fig.~\ref{fig:file_structure}.
This python file takes the model name as an input argument and creates a new model directory containing model and preprocessing files using that name.
The structure of this new model directory will match those of pre-existing models like I3D, C3D, ResNet50+LSTM, and TSN, as seen in the center of Fig.~\ref{fig:file_structure}.

The majority of code development occurs within the model and preprocessing files.
The model file contains \texttt{TODO} statements in the methods ``\texttt{inference}'' and ``\texttt{loss}'' which must be implemented by the user with the available layer definitions and any Tensorflow loss function.
Other optional functions, for example \texttt{preprocess\_tfrecords} and \texttt{load\_default\_weights}, are available and have default initializations.

Within the preprocessing file, a Tensorflow tensor containing a clip is passed as input.
The expectation is that the final tensor that is returned, after preprocessing and reshaping the input, has a consistent shape in terms of frame count, height, width, and number of channels.
Once a valid model and its functions have been defined, it can immediately be trained and tested upon, using \texttt{train.py} and \texttt{test.py} respectively.


\section{Benchmarks}
\label{sec:benchmarks}

A main contribution of \acro~is the access it provides to activity classification results on SOTA models under a common platform.
In this section, we provide numerical evaluations of \acro~in terms of the activity classification performance of available SOTA models as well as speed and memory capacity.

\subsection{SOTA Performance}
Activity classification models can be roughly divided into two categories: 3D convolution-based and 2D convolution-based that apply a temporal consensus.
Of the four SOTA models that have been implemented in \acro, two are 3D convolution-based models, I3D~\cite{carreira2017quo} and C3D~\cite{tran2015learning}, and two are 2D convolution-based models, TSN~\cite{wang2016temporal} and ResNet50+LSTM~\cite{he2016deep}.
The focus of \acro~is currently on RGB-based models only, thus two stream networks~\cite{simonyan2014two} have not been implemented.

HMDB51~\cite{kuehne2013hmdb51} and UCF101~\cite{soomro2012ucf101} are the activity classification datasets that have been used to replicate the performance of the SOTA models.
These datasets were selected because the original authors of all four of the SOTA models provide results on both datasets.
Newer and larger activity classification datasets, for example Kinetics~\cite{he2016deep}, or even event classification datasets, for example Moments in Time~\cite{monfortmoments}, can be converted to TFRecords format and used in \acro.
Table \ref{tab:model_results} shows the expected \acro~ performance of the four models I3D, C3D, TSN, and ResNet50+LSTM on HMDB51 and UCF101 and compares this performance with that released by the original authors.
The deviation in performances between the original and \acro~models can be attributed to a number of reasons including the lack of public model weights, missing details in the preprocessing and training procedures, unspecified hyperparameters during fine-tuning, and a conversion from the model's native deep learning library to Tensorflow.
The following subsections will detail the implementation of each model.


\subsubsection{I3D}
\label{subsubsec:i3d}
I3D~\cite{carreira2017quo} has the highest published performance out of all models implemented in \acro.
It uses the Inception-V1~\cite{szegedy2015going} network as a backbone and replicates each 2D convolutional filter to form 3D convolutions.
While the code released for I3D is written in Tensorflow, only model weights trained on the Kinetics dataset~\cite{he2016deep} and a testing script are provided.
Since no training code has been released, the I3D training procedure in \acro~is reproduced from the written descriptions of the training process in the code repository and the original paper itself.
Due to the uncertainty in the training procedure, there exists a drop in performance between the \acro~and the authors results on HMDB51 and UCF101.
However, even with the drop in performance I3D continues to be the best performing model in \acro.

\subsubsection{C3D}
\label{subsubsec:c3d}
C3D~\cite{tran2015learning} is built around 3D convolutions, like I3D, and is commonly used for feature extraction and fine-tuning on video-based tasks.
The original implementation of C3D is publicly available along with the fine-tuneing procedure on UCF101.
However, all code is written using Caffe and only the weights for C3D trained on Sports-1M~\cite{KarpathyCVPR14} are given.
The \acro~implementation of the model and its training and testing pipelines follow the original Caffe code, used to fine-tune the model on UCF101, as closely as possible.

In the work introducing C3D~\cite{tran2015learning}, the model trained on Sports-1M is tested on UCF101 by extracting features and training an SVM.
However, in subsequent work~\cite{tran2017convnet} C3D has been fine-tuned on both UCF101 and HMDB51.
We fine-tune and test C3D, rather than train an SVM, to maintain consistency with the other models in \acro.

\subsubsection{TSN}
\label{subsubsec:tsn}
The original TSN~\cite{wang2016temporal} model is written using Caffe, though the authors also provide an implementation in Pytorch.
TSN is the only model of those in \acro~where the original authors have provided both HMDB51 and UCF101 fine-tuned weights.
In an effort to replicate the published performance as closely as possible, we use the provided weights in the presented results.
However, the training pipeline for TSN has also been reimplemented in \acro~so that the model can be retrained if desired.

\subsubsection{ResNet50 + LSTM}
\label{subsubsec:resnet}
The original concept of training an LSTM on the features from a 2D convolutional neural network (CNN) was concieved in the LRCN model~\cite{donahue2015long}.
ResNet50 + LSTM is a baseline presented in conjunction with the Kinetics~\cite{he2016deep} dataset and is used in \acro~because it achieves a higher performance than LRCN.
The model was constructed, trained, and tested according to the descriptions given in the original paper~\cite{he2016deep} since no code or weights are available for it.
While this model may not perform as well as other models, LSTM~\cite{hochreiter1997long}-based models are common in video tasks and should exist in a video classification platform.

\subsection{Memory and Speed Tests}

The pace of growth in hardware memory size is smaller than the growth in network dimensionality making it crucial that \acro~maintains as small a memory footprint as possible when training and testing models.
\acro~is designed to be compatible with a CUDA-ready Nvidia GPU with sufficient VRAM to load a model with at least a single mini-batch of data into memory.

Table~\ref{tab:speed} shows the amount of GPU memory that each model within \acro~uses when the mini-batch size is set to one.
I3D and C3D consume the most memory due to their use of 3D convolutional filters.
ResNet50+LSTM uses significantly less since it only consists of a 2D CNN feeding into an LSTM.
TSN is much smaller than any other model since it consists of only a 2D CNN classifying frames individually and averaging the predictions.
Because of this, the mini-batch size during training can be much larger for TSN than for any other model.

Tensorflow was selected as the backbone of \acro, in part, due to the performance and minimal overhead that can be achieved during runtime.
Table~\ref{tab:speed} enumerates the results from performance tests and quantifies the time taken to train networks and calculate classification metrics within \acro.
For most models, the difference between training and testing speed is less than $0.25$ seconds. 
However, the testing speed of TSN is nearly five times its training speed.
This is likely due to the use of oversampling during the preprocessing of testing videos for TSN.
Oversampling a video creates ten copies of each frame where each copy is either a crop, either from the center or a corner of the frame, or the mirrored versions of a crop.

The data shown in Table~\ref{tab:speed} should only be taken as a reference of the performance achieved in \acro.
The speed and memory consumption of \acro~will likely vary between users due to differing hardware and environments.

\begin{table}[t!]
    \begin{center}
        \begin{tabular}{c|c|c|c}
            \hline
            Model & Memory & Train Speed & Test Speed \\
            \hline
            I3D &  8.46 Gb & 0.91 s & 1.15 s  \\
            C3D & 7.44 Gb & 0.17 s & 0.14 s \\
            TSN & 0.91 Gb & 0.18 s & 0.92 s \\
            ResNet50+LSTM & 4.36 Gb & 0.84 s & 0.93 s \\
            \hline
        \end{tabular}
    \end{center}
        \caption{The table shows the GPU memory, shown in gigabytes, used by four SOTA activity classification models in \acro.
        These values were obtained when training on HMDB51 on a NVIDIA GTX 1080ti video card using a mini-batch size of one.
        The table also shows the speed, in seconds, to train and test these models. 
        The training and testing times, on HMDB51 with a mini-batch size of one, are averaged over 1000 iterations.
        }
        \label{tab:speed}
\end{table}


\section{Conclusion}
\label{sec:conclusion}

In field of activity classification, SOTA models are spread across languages and research pipelines making it difficult to compare and build upon their results.
For example, the initial weights of C3D and TSN were trained in Caffe, ResNet50 in Keras, and I3D in Tensorflow.
Each was written and trained using different hyperparameters which are often not available to the public.
In these cases, experimentation is required to replicate the performance of a model.
We present \acro, an open-source Tensorflow-based platform which provides a unified pipeline with access to SOTA activity classification models and an environment to develop new models.

\acro~will serve as a basis for future models and remove the burden of replicating results and reimplementing networks from the community.
\acro~houses a variety of blocks and modules which make it quick and easy for a user to generate results on a model or develop their own model.
The training and testing of these models has been streamlined down to command line calls with multiple arguments which can be used to alter the flow of data.
Through GitHub\footnote{\url{https://www.github.com/MichiganCOG/M-PACT}}, \acro~will accept pull requests of new activity classification methods that are developed in the platform.
The goal of \acro~is to serve as the all encompassing and continually updated benchmarking and development platform for SOTA activity classification.




\section{Acknowledgements}

Supported by the Intelligence Advanced Research Projects Activity (IARPA) via Department of Interior/ Interior Business Center (DOI/IBC) contract number D17PC00341. The U.S. Government is authorized to reproduce and distribute reprints for Governmental purposes notwithstanding any copyright annotation thereon.  Disclaimer:  The  views  and  conclusions  contained  herein  are  those  of the authors and should not be interpreted as necessarily representing the official policies or endorsements, either expressed or implied, of IARPA, DOI/IBC, or the U.S. Government.

\newpage
{\small
\bibliographystyle{ieee}
\bibliography{egbib}
}

\end{document}